\PassOptionsToPackage{usenames}{color}
\pdfoutput=1 
\documentclass[11pt,a4paper]{article}

\usepackage{relsize} 
\usepackage[hyperref]{style/emnlp2017}

\usepackage{microtype}
\usepackage{ dsfont }
\usepackage[boxed]{algorithm2e}

\usepackage[small,bf,skip=5pt]{caption}
\usepackage{sidecap} 
\usepackage{rotating}	

\usepackage{titlesec}
\titleformat*{\subparagraph}{\itshape}
\titlespacing{\subparagraph}{%
  1em}{
  0pt}{
  1em}

\usepackage{lingmacros}

\usepackage[shortlabels]{enumitem} 
\setitemize{noitemsep,topsep=0em,leftmargin=*}
\setenumerate{noitemsep,leftmargin=0em,itemindent=13pt,topsep=0em}

\usepackage{xspace}
\usepackage{xparse} 

\usepackage{textcomp}

\usepackage{framed}

\usepackage{listings}

\usepackage{amssymb}	
\usepackage{amsmath}

\usepackage{mathptmx}	
\usepackage[scaled=.8]{beramono}
\usepackage[scaled=.85]{helvet}
\usepackage[T1]{fontenc}
\usepackage[utf8x]{inputenc}

\usepackage{MnSymbol}	

\usepackage{latexsym}

\usepackage{array}
\usepackage{multirow}
\usepackage{booktabs} 
\usepackage{multicol}
\usepackage{footnote}
\newcolumntype{H}{>{\setbox0=\hbox\bgroup}c<{\egroup}@{}} 

\usepackage[usenames]{color}
\usepackage{xcolor}

\usepackage[normalem]{ulem} 
\usepackage{colortbl}
\usepackage{graphicx}
\usepackage{subcaption}
\usepackage{tikz}
\usetikzlibrary{arrows,positioning,calc}

\emnlpfinalcopy 

\usepackage[multiple]{footmisc}

\usepackage{nameref}
\usepackage{cleveref}

\crefformat{part}{\S#2#1#3}
\crefformat{chapter}{\S#2#1#3}
\crefformat{section}{\S#2#1#3}
\crefformat{subsection}{\S#2#1#3}
\crefformat{subsubsection}{\S#2#1#3}
\crefformat{paragraph}{\P#2#1#3}
\crefformat{subparagraph}{\P#2#1#3}
\crefmultiformat{section}{\S#2#1#3}{ and~\S#2#1#3}{, \S#2#1#3}{, and~\S#2#1#3}
\crefmultiformat{subsection}{\S#2#1#3}{ and~\S#2#1#3}{, \S#2#1#3}{, and~\S#2#1#3}
\crefmultiformat{subsubsection}{\S#2#1#3}{ and~\S#2#1#3}{, \S#2#1#3}{, and~\S#2#1#3}
\crefmultiformat{paragraph}{\P\P#2#1#3}{ and~#2#1#3}{, #2#1#3}{, and~#2#1#3}
\crefmultiformat{subparagraph}{\P\P#2#1#3}{ and~#2#1#3}{, #2#1#3}{, and~#2#1#3}
\crefrangeformat{section}{\mbox{\S\S#3#1#4--#5#2#6}}
\crefrangeformat{subsection}{\mbox{\S\S#3#1#4--#5#2#6}}
\crefrangeformat{subsubsection}{\mbox{\S\S#3#1#4--#5#2#6}}
\crefrangeformat{paragraph}{\mbox{\P\P#3#1#4--#5#2#6}}
\crefrangeformat{subparagraph}{\mbox{\P\P#3#1#4--#5#2#6}}
\crefname{part}{Part}{Parts}
\Crefname{part}{Part}{Parts}
\crefname{chapter}{ch.}{ch.}
\Crefname{chapter}{Ch.}{Ch.}
\crefname{figure}{figure}{figures}
\crefname{subfigure}{figure}{figures}
\Crefname{subfigure}{Figure}{Figures}
\crefname{appsec}{appendix}{appendices}
\Crefname{appsec}{Appendix}{Appendices}
\crefname{algocf}{algorithm}{algorithms}
\Crefname{algocf}{Algorithm}{Algorithms}
\crefname{enums,enumsi}{example}{examples}
\Crefname{enums,enumsi}{Example}{Examples}
\crefname{}{example}{examples} 
\Crefname{}{Example}{Examples}
\crefformat{enums}{(#2#1#3)}
\crefformat{enumsi}{(#2#1#3)}
\crefformat{}{(#2#1#3)}
\crefrangeformat{enums}{\mbox{(#3#1#4--#5#2#6)}}
\crefrangeformat{enumsi}{\mbox{(#3#1#4--#5#2#6)}}

\ifx\creflastconjunction\undefined%
\newcommand{\creflastconjunction}{, and\nobreakspace} 
\else%
\renewcommand{\creflastconjunction}{, and\nobreakspace} 
\fi%

\newcommand*{\Fullref}[1]{\hyperref[{#1}]{\Cref*{#1}: \nameref*{#1}}}
\newcommand*{\fullref}[1]{\hyperref[{#1}]{\cref*{#1}: \nameref{#1}}}

\NewDocumentEnvironment{itmize}{}{\begin{itemize}[noitemsep]}{\end{itemize}}
\NewDocumentEnvironment{enumrate}{}{\begin{enumerate}[noitemsep]}{\end{enumerate}}
\let\Item\item
\renewcommand\enddescription{\endlist\global\let\item\Item}
\NewDocumentEnvironment{describe}{}{\renewcommand\item[1][]{\Item \textbf{##1:} }\begin{itemize}}{\end{itemize}}
\NewDocumentEnvironment{edescribe}{}{\renewcommand\item[1][]{\Item \textbf{##1:} }\begin{enumerate}}{\end{enumerate}}

\usepackage{color}
\usepackage{bm}
\definecolor{orange}{rgb}{1,0.5,0}
\definecolor{mdgreen}{rgb}{0.05,0.6,0.05}
\definecolor{mdblue}{rgb}{0,0,0.7}
\definecolor{dkblue}{rgb}{0,0,0.5}
\definecolor{dkgray}{rgb}{0.3,0.3,0.3}
\definecolor{slate}{rgb}{0.25,0.25,0.4}
\definecolor{gray}{rgb}{0.5,0.5,0.5}
\definecolor{ltgray}{rgb}{0.7,0.7,0.7}
\definecolor{purple}{rgb}{0.7,0,1.0}
\definecolor{lavender}{rgb}{0.65,0.55,1.0}

\lstnewenvironment{Python}[1][]
  {\lstset{language=Python,
           morekeywords=as,
           #1}%
  }
  {}
  
\lstnewenvironment{Output}[1][]
  {\lstset{#1}%
  }
  {}

\makeatletter
\renewcommand{\paragraph}{%
  \@startsection{paragraph}{4}%
  {\z@}{.2ex \@plus 1ex \@minus .2ex}{-1em}%
  {\normalfont\normalsize\bfseries}%
}
\makeatother

\newcommand{\fnf}[1]{\textsc{\textsf{#1}}} 
\newcommand{\fnr}[1]{\textbf{\textsf{#1}}} 
\newcommand{\fnrel}[1]{\textsl{#1}} 
\newcommand{\fnst}[1]{\textsl{#1}} 

\newcommand{\lex}[1]{\textit{#1}} 

\newcommand{\srsversion}[1]{}
\newcommand{\finalversion}[1]{#1}

\newcommand{\longversion}[1]{} 
\newcommand{\subversion}[1]{∏} 
\newcommand{\draftnotice}[1]{} 
\newcommand{\nonanonversion}[1]{} 

\title{The NLTK FrameNet API:\\ Designing for Discoverability with a Rich Linguistic Resource}

\author{Nathan Schneider \\
Georgetown University \\
Washington, DC \\
{\tt nathan.schneider@georgetown.edu}
\And
Chuck Wooters \\
Semantic Machines\\
Berkeley, CA\\
{\tt wooters@semanticmachines.com}}

\begin{document}
\maketitle
\begin{abstract}
A new Python API, integrated within the NLTK suite, 
offers access to the FrameNet 1.7 lexical database. 
The lexicon (structured in terms of frames) as well as annotated sentences 
can be processed programatically, or browsed with human-readable displays 
via the interactive Python prompt. 
\end{abstract}

\section{Introduction}

For over a decade, the Berkeley FrameNet (henceforth, simply ``FrameNet'') project \citep{baker-98} 
has been documenting the vocabulary of contemporary English with respect to the theory of 
frame semantics \citep{fillmore-82}. A freely available, linguistically-rich 
resource, FrameNet now covers over 1,000 semantic frames, 
10,000 lexical senses, and 100,000 lexical annotations in sentences drawn from corpora.
The resource has formed a basis for much research in natural language processing---%
most notably, a tradition of semantic role labeling that continues to this day \citep[\emph{inter alia}]{gildea-02,baker-07,das-14,fitzgerald-15,roth-15}.

Despite the importance of FrameNet, computational users are often frustrated 
by the complexity of its custom XML format. Whereas much of the resource 
is browsable on the web (\url{http://framenet.icsi.berkeley.edu/}), 
certain details of the linguistic descriptions and annotations 
languish in obscurity as they are not exposed by the HTML views of the data.\footnote{For example, 
one of the authors was recently asked by a FrameNet user whether 
frame-to-frame relations include mappings between individual frame elements. 
They do, but the user's confusion is understandable because 
these mappings are not exposed in the HTML frame definitions on 
the website. (They can be explored visually via the FrameGrapher tool on the website, 
\url{https://framenet.icsi.berkeley.edu/fndrupal/FrameGrapher}, if the user 
knows to look there.) In the interest of space, our API does not show them in the frame display, 
but they can be accessed via an individual frame relation object or with the \texttt{fe\_relations()} method, \cref{sec:lexicon}.}
The few open source APIs for reading FrameNet data are now antiquated, 
and none has been widely adopted.\footnote{We are aware of:
\begin{itemize}
  \item \href{https://github.com/dasmith/FrameNet-python}{\tt github.com/dasmith/FrameNet-python} 
(Python)
\item \href{https://nlp.stanford.edu/software/framenet.shtml}{\tt {nlp.stanford.edu/software/framenet.shtml}} (Java)
\item \href{https://github.com/FabianFriedrich/Text2Process/tree/master/src/de/saar/coli/salsa/reiter/framenet}{\nolinkurl{github.com/FabianFriedrich/Text2Process/tree/master/src/de/saar/coli/salsa/reiter/framenet}} (Java)
\item \href{https://github.com/GrammaticalFramework/gf-contrib/tree/master/framenet}{\nolinkurl{github.com/GrammaticalFramework/gf-contrib/tree/master/framenet}} (Grammatical Framework)
\end{itemize}
None of these has been updated in the past few years, 
so they are likely not fully compatible with the latest data release.}

We describe a new, user-friendly Python API for accessing FrameNet data.
The API is included within recent releases of the popular NLTK suite \citep{nltk},
and provides access to nearly all the information in the FrameNet release.

\section{Installation}

Instructions for installing NLTK are found at \mbox{\url{nltk.org}}.
NLTK is cross-platform and supports Python 2.7 as well as Python 3.x environments. 
It is bundled in the Anaconda and Enthought Canopy 
Python distributions for data scientists.\footnote{\url{https://www.continuum.io/downloads},\\ \url{https://store.enthought.com/downloads}}

In a working NLTK installation (version 3.2.2 or later), 
one merely has to invoke a method to download the 
FrameNet data:\footnote{{\tt >{}>{}>} 
is the standard Python interactive prompt, 
generally invoked by typing {\tt python} on the command line. 
Python code can then be entered at the prompt, where it is evaluated/executed. 
Henceforth, examples will assume familiarity with the basics of Python.}\footnote{By default, 
the 855MB data release is installed under the user's home directory, but an alternate location can be specified: 
see \url{http://www.nltk.org/data.html}.}

\begin{Python}
>>> import nltk
>>> nltk.download('framenet_v17')
\end{Python}

Subsequently, the {\tt framenet} module is loaded as follows 
(with alias {\tt fn} for convenience):

\begin{Python}
>>> from nltk.corpus import framenet as fn
\end{Python}

\section{Overview of FrameNet}

FrameNet is organized around conceptual structures known as \textbf{frames}. 
A semantic frame represents a \textbf{scene}---a kind of event, state, or other scenario 
which may be universal or culturally-specific, and domain-general or domain-specific. 
The frame defines participant roles or \textbf{frame elements} (FEs), whose relationships 
forms the conceptual background required to understand 
(certain senses of) vocabulary items. 
Oft-cited examples by Fillmore include:
\begin{itemize}
  \item Verbs such as \lex{buy}, \lex{sell}, and \lex{pay},
  and nouns such as \lex{buyer}, \lex{seller}, \lex{price}, and \lex{purchase}, 
  are all defined with respect to a commercial transaction scene (frame). 
  FEs that are central to this frame---they may or may not be mentioned explicitly 
  in a text with one of the aforementioned lexical items---are the \fnr{Buyer}, 
  the \fnr{Seller}, the \fnr{Goods} being sold by the \fnr{Seller}, 
  and the \fnr{Money} given as payment in exchange by the \fnr{Buyer}.
  \item The concept of \fnf{Revenge}---lexicalized in vocabulary items such as 
  \lex{revenge}, \lex{avenge}, \lex{avenger}, \lex{retaliate}, \lex{payback}, 
  and \lex{get even}---fundamentally presupposes an \fnr{Injury} that an 
  \fnr{Offender} has inflicted upon an \fnr{Injured\_party}, for which an 
  \fnr{Avenger} (who may or may not be the same as the \fnr{Injured\_party}) 
  seeks to exact some \fnr{Punishment} on the \fnr{Offender}.
  \item A \lex{hypotenuse} presupposes a geometrical notion of right triangle, 
  while a \lex{pedestrian} presupposes a street with both vehicular and nonvehicular 
  traffic. (Neither frame is currently present in FrameNet.)
\end{itemize}
The FEs in a frame are formally listed alongside an English description 
of their function within the frame.
Frames are organized in a network, including an inheritance hierarchy 
(e.g., \fnf{Revenge} is a special case of an \fnf{Event}) and other kinds of 
frame-to-frame relations.

Vocabulary items listed within a frame are called \textbf{lexical units} (LUs).
FrameNet's inventory of LUs includes both content and function words.
Formally, an LU links a lemma with a frame.\footnote{The lemma name incorporates
a part-of-speech tag. The lemma may consist of a single 
word, such as \lex{surrender.v}, or multiple words, such as \lex{give up.v}.}

In a text, a token of an LU is said to \textbf{evoke} the frame. 
Sentences are annotated with respect to frame-evoking tokens 
and their FE spans. Thus: 
\begin{center}
[Snape]$_{\fnr{Injured\_party}}$ 's \textbf{revenge} [on Harry]$_{\fnr{Offender}}$
\end{center}
labels overt mentions of participants in the \fnf{Revenge} frame.

The reader is referred to \citep{framenet} for a contemporary 
introduction to the resource and the theory of frame semantics 
upon which it is based. Extensive linguistic details 
are provided in \citep{ruppenhofer-16}.

\section{API Overview}

\subsection{Design Principles}

The API is designed with the following goals in mind:

\paragraph{Simplicity.} It should be easy to access important objects in the database 
(primarily frames, lexical units, and annotations), whether by 
iterating over all entries or searching for particular ones.
To avoid cluttering the API with too many methods, other information in the database 
should be reachable via object attributes. 
Calling the API's \texttt{help()} method prints a summary of the main methods 
for accessing information in the database.

\paragraph{Discoverability.} Many of the details of the database are complex. 
The API makes it easy to browse what is in database objects via the Python 
interactive prompt. The main way it achieves this is with pretty-printed 
displays of the objects, such as the frame display in \cref{fig:revengedisplay} 
(see \cref{sec:objs}). The display makes it clear how to access attributes 
of the object that a novice user of FrameNet might not have known about.

In our view, this approach sets this API apart from others. 
Some of the other NLTK APIs for complex structured data make it difficult to browse 
the structure without consulting documentation.

\paragraph{On-demand loading.} The database is stored in thousands of XML files, 
including files indexing the lists of frames, frame relations, LUs, and full-text documents, 
plus individual files for all frames, LUs, and full-text documents.
Unzipped, the FrameNet 1.7 release is 855 MB. 
Loading all of these files---particularly the corpus annotations---is slow and memory-intensive, 
costs which are unnecessary for many purposes.
Therefore, the API is carefully designed with lazy data structures to load 
XML files only as needed. Once loaded, all data is cached in memory for fast subsequent access.


\begin{figure*}[t]\small
\begin{Output}
frame (347): Revenge

[URL] https://framenet2.icsi.berkeley.edu/fnReports/data/frame/Revenge.xml

[definition]
  This frame concerns the infliction of punishment in return for a
  wrong suffered. An Avenger performs a Punishment on a Offender as
  a consequence of an earlier action by the Offender, the Injury.
  The Avenger inflicting thePunishment need not be the same as the
  Injured_Party who suffered the Injury,  but the Avenger does have
  to share the judgment that the Offender's action was wrong. The
  judgment that the Offender had inflicted an Injury  is made
  without regard to the law.  '(1) They took revenge for the deaths
  of two loyalist prisoners.' '(2) Lachlan went out to avenge
  them.' '(3) The next day, the Roman forces took revenge on their
  enemies..'

[semTypes] 0 semantic types

[frameRelations] 1 frame relations
  <Parent=Rewards_and_punishments -- Inheritance -> Child=Revenge>

[lexUnit] 18 lexical units
  avenge.v (6056), avenger.n (6057), get back (at).v (10003), get
  even.v (6075), payback.n (10124), retaliate.v (6065),
  retaliation.n (6071), retribution.n (6070), retributive.a (6074),
  retributory.a (6076), revenge.n (6067), revenge.v (6066),
  revengeful.a (6073), revenger.n (6072), sanction.n (10676),
  vengeance.n (6058), vengeful.a (6068), vindictive.a (6069)


[FE] 14 frame elements
            Core: Avenger (3009), Injured_party (3022), Injury (3018), 
Offender (3012), Punishment (3015)
      Peripheral: Degree (3010), Duration (12060), Instrument (3013), 
Manner (3014), Place (3016), Purpose (3017), Time (3021)
  Extra-Thematic: Depictive (3011), Result (3020)

[FEcoreSets] 2 frame element core sets
  Injury, Injured_party
  Avenger, Punishment
\end{Output}
\caption{Textual display of the \fnf{Revenge} frame. 
Shown in square brackets are attribute names for accessing the frame's contents.
In parentheses are IDs for the frame, its LUs, and its FEs.}
\label{fig:revengedisplay}
\end{figure*}

\subsection{Lexicon Access Methods}

The main methods for looking up information in the lexicon are:
\vspace{2mm}
\begin{table}[h]\centering
\begin{tabular}{>{\texttt\bgroup}l<{\egroup}>{\texttt\bgroup}l<{\egroup}}
\textbf{frames}(name) & \textbf{frame}(nameOrId) \\
\textbf{lus}(name, frame)    & \textbf{lu}(id) \\
\multicolumn{2}{>{\texttt\bgroup}l<{\egroup}}{\textbf{fes}(name, frame)} \\
\end{tabular}
\end{table}

\noindent The methods with plural names (left) are for searching the lexicon 
by regular expression pattern to be matched against the entry name. 
In addition (or instead), \texttt{lus()} and \texttt{fes()} allow for the results 
to be restricted to a particular frame. 
The result is a list with 0 or more elements.
If no arguments are provided, all entries in the lexicon are returned. 

An example of a search by 
frame name pattern:\footnote{\texttt{(?i)} makes the pattern case-insensitive.}
\begin{Python}
>>> fn.frames('(?i)creat')
[<frame ID=268 name=Cooking_creation>, 
 <frame ID=1658 name=Create_physical_artwork>, ...]
\end{Python}

Similarly, a search by LU name pattern---note that the \texttt{.v} suffix 
is used for all verbal LUs:
\begin{Python}
>>> fn.lus(r'.+en\.v')
[<lu ID=5331 name=awaken.v>, 
 <lu ID=7544 name=betoken.v>, ...]
\end{Python}

The \texttt{frame()} and \texttt{lu()} methods are for retrieving a single known entry by its name or ID.
Attempting to retrieve a nonexistent entry triggers an exception of type \texttt{FramenetError}.

Two additional methods are available for frame lookup:
\texttt{frame\_ids\_and\_names(name)} to get a mapping from frame IDs to names,
and \texttt{frames\_by\_lemma(name)} to get all frames with some LU matching the given name pattern.

\begin{figure*}\small
\begin{Output}
exemplar sentence (929548):
[sentNo] 0
[aPos] 1113164

[LU] (6067) revenge.n in Revenge

[frame] (347) Revenge

[annotationSet] 2 annotation sets

[POS] 12 tags

[POS_tagset] BNC

[GF] 4 relations

[PT] 4 phrases

[text] + [Target] + [FE] + [Noun]

A short while later Joseph had his revenge on Watney 's . 
------------------- ------ ^^^ --- ******* ------------
Time                Avenge sup Ave         Offender     
 
 
 
[Injury:DNI]
 (Avenge=Avenger, sup=supp, Ave=Avenger) 
\end{Output}
\caption{A lexicographic sentence display. The visualization of the frame annotation set at the bottom is produced 
by pretty-printing the combined information in the \texttt{text}, \texttt{Target}, \texttt{FE}, and \texttt{Noun} layers. 
Abbreviations in the visualization are expanded at the bottom in parentheses (``supp'' is short for ``support''). 
``DNI'' is FrameNet jargon for ``definite null instantiation''; \texttt{GF} stands for ``grammatical function''; 
and \texttt{PT} stands for ``phrase type''.}
\label{fig:revengesent}
\end{figure*}

\subsection{Database Objects}\label{sec:objs}

All structured objects in the database---frames, LUs, FEs, etc.---are loaded  
as \texttt{AttrDict} data structures. Each \texttt{AttrDict} instance is a mapping 
from string keys to values, which can be strings, numbers, or structured objects. 
\texttt{AttrDict} is so called because it allows keys to be accessed as attributes:

\begin{Python}
>>> f = fn.frame('Revenge')
>>> f.keys()
dict_keys(['cBy', 'cDate', 'name', 'ID', '_type', 
'definition', 'definitionMarkup', 'frameRelations', 
'FE', 'FEcoreSets', 'lexUnit', 'semTypes', 'URL'])
>>> f.name
'Revenge'
>>> f.ID
347
\end{Python}

For the most important kinds of structured objects, the API specifies  
textual \textbf{displays} that organize the object's contents 
in a human-readable fashion. \Cref{fig:revengedisplay} shows the display 
for the \fnf{Revenge} frame, which would be printed by entering 
\lstinline|fn.frame('Revenge')| at the interactive prompt.
The display gives attribute names in square brackets; e.g., 
\lstinline|lexUnit|, which is a mapping from LU names to objects. 
Thus, after the code listing in the previous paragraph, 
\lstinline|f.lexUnit['revenge.n']|
would access to one of the LU objects in the frame, 
which in turn has its own attributes and textual display.

\begin{figure*}[t]
\begin{Output}
full-text sentence (4148528) in Tiger_Of_San_Pedro:


[POS] 25 tags

[POS_tagset] PENN

[text] + [annotationSet]

They 've been looking for him all the time for their revenge , 
              *******                                *******   
              Seeking                                Revenge   
              [3] ?                                  [2]       

but it is only now that they have begun to find him out . "
                                  *****    ****
                                  Proce    Beco
                                  [1]      [4] 
 (Proce=Process_start, Beco=Becoming_aware)
\end{Output}
\caption{A sentence of full-text annotation. If this sentence object is stored 
under the variable \lstinline{sent}, its frame annotation with respect to the 
target \textit{revenge} is accessed as \lstinline{sent.annotationSet[2]}. 
(The \texttt{?} under \textit{looking} indicates that there is no corresponding 
LU defined in the \fnf{Seeking} frame; in some cases the full-text annotators 
marked but did not define out-of-vocabulary LUs which fit an existing frame. 
Also, some full-text annotation sets annotate an LU without its FEs---these 
are shown with \texttt{!} to reflect the annotation set's status code of \texttt{UNANN}.)
}
\label{fig:ftsent}
\end{figure*}

\subsection{Advanced Lexicon Access}\label{sec:lexicon}

\paragraph{Frame relations.} The inventory of frames is organized in a semantic network 
via several kinds of frame-to-frame relations. 
For instance, the \fnf{Revenge} frame is involved in one frame-to-frame relation:
it is related to the more general \fnf{Rewards\_and\_punishments} frame by \fnrel{Inheritance}, 
as shown in the middle of \cref{fig:revengedisplay}. \fnf{Rewards\_and\_punishments}, 
in turn, is involved in \fnrel{Inheritance} relations with other frames. 
Each frame-to-frame relation bundles mappings between corresponding FEs in the two frames.

Apart from the \texttt{frameRelations} attribute of frame objects, 
frame-to-frame relations can be browsed by the main method 
\texttt{frame\_relations(frame, frame2, type)}, where the optional arguments 
allow for filtering by one or both frames and the kind of relation. 
Within a frame relation object, pairwise FE relations are stored in the 
\texttt{feRelations} attribute. Main method \texttt{fe\_relations()} 
provides direct access to links between FEs.
The inventory of relation types, including \fnrel{Inheritance}, \fnrel{Causative}, \fnrel{Inchoative}, 
\fnrel{Subframe}, \fnrel{Perspective\_on}, and others, is available via main method 
\texttt{frame\_relation\_types()}.

\paragraph{Semantic types.} These provide additional semantic categorizations 
of FEs, frames, and LUs. For FEs, they mark selectional restrictions 
(e.g., \lstinline{f.FE['Avenger'].semType} gives the \fnst{Sentient} type).
Main method \texttt{propagate\_semtypes()} propogates the FE semantic type labels 
marked explicitly to other FEs according to inference rules that follow the FE relations. 
This should be called prior to inspecting FE semtypes (it is not called by default 
because it takes several seconds to run).

The semantic types are database objects in their own right, 
and they are organized in their own inheritance hierarchy.
Main method \texttt{semtypes()} returns all semantic types as a list;
main method \texttt{semtype()} looks up a particular one by name, ID, or abbreviation; 
and main method \texttt{semtype\_inherits()} checks whether two semantic types have a 
subtype--supertype relationship.


\subsection{Corpus Access}

Frame-semantic annotations of sentences can be accessed via 
the \texttt{exemplars} and \texttt{subCorpus} attributes of an LU object, 
or via the following main methods:
\vspace{1mm}
\begin{table}[h]\centering
\begin{tabular}{@{}>{\texttt\bgroup}l<{\egroup}@{~}>{\texttt\bgroup}l<{\egroup}@{~}>{\texttt\bgroup}l<{\egroup}@{}}
\multicolumn{3}{@{}>{\texttt\bgroup}l<{\egroup}@{}}{\textbf{annotations}(luname, exemplars, full\_text)} \\
\textbf{sents}()    & \textbf{exemplars}(luname) & \textbf{ft\_sents}(docname) \\
\textbf{doc}(id)    & \textbf{docs}(name) 
 \\
\end{tabular}
\end{table}

\texttt{annotations()} returns a list of frame \textbf{annotation sets}. 
Each annotation set consists of a frame-evoking \textbf{target} (token) within a sentence, 
the LU in the frame it evokes, its overt FE spans in the sentence, 
and the status of null-instatiated FEs.\footnote{In frame semantics, core FEs that are not overt
but are conceptually required by a frame are said to be implicit via null instantiation \citep{framenet}.}
Optionally, the user may filter by LU name, or limit by the type of annotation (see next paragraph):
\texttt{exemplars} and \texttt{full\_text} both default to \texttt{True}.
In the XML, the components of an annotation set are stored in several 
annotation layers: one (and sometimes more than one) layer of FEs, 
as well as additional layers for other syntactic information 
(including grammatical function and phrase type labels for each FE, 
and copular or support words relative to the frame-evoking target). 

Annotation sets are organized by sentence. Corpus sentences appear in two kinds of annotation: 
\texttt{exemplars()} retrieves sentences with lexicographic annotation 
(where a single target has been selected for annotation to serve as an example of an LU); 
the optional argument allows for filtering the set of LUs. 
\texttt{ft\_sents()} retrieves sentences from documents selected for full-text annotation 
(as many targets in the document as possible have been annotated); 
the optional argument allows for filtering by document name.
\texttt{sents()} can be used to iterate over all sentences.
Technically, each sentence object contains multiple annotation sets: 
the first is for sentence-level annotations, including 
the part-of-speech tagging and in some cases named entity labels;
subsequent annotation sets are for frame annotations. 
As lexicographic annotations have only one frame annotation set, it is visualized in the sentence display: 
\cref{fig:revengesent} shows the display for \lstinline|f.lexUnit['revenge.n'].exemplars[20]|.
Full-text annotations display target information only, 
allowing the user to drill down to see each annotation set, 
as in \cref{fig:ftsent}.

Sentences of full-text annotation can also be browsed by document using the 
\texttt{doc()} and \texttt{docs()} methods. 
The document display lists the sentences with numeric offsets.

\section{Limitations and future work}

The main part of the Berkeley FrameNet data that the API currently does \emph{not} support 
are \textbf{valence patterns}. For a given LU, the valence patterns summarize
the FEs' syntactic realizations across annotated tokens. 
They are displayed in each LU's ``Lexical Entry'' report on the FrameNet website.

We intend to add support for valence patterns in future releases, along with 
more sophisticated querying\slash browsing capabilities for annotations,
and better displays for syntactic information associated with FE annotations. 
Some of this functionality can be modeled after tools like FrameSQL \citep{framesql} 
and Valencer \citep{valencer}.
In addition, it is worth investigating whether the API can be adapted for 
FrameNets in other languages, and to support cross-lingual mappings being added 
to 14 of these other FrameNets in the ongoing Multilingual FrameNet project.\footnote{\href{https://github.com/icsi-berkeley/multilingual_FN}{\tt github.com/icsi-berkeley/multilingual\_FN}}

%

\finalversion{\section*{Acknowledgments}
We thank Collin Baker, Michael Ellsworth, and Miriam R.~L. Petruck 
for helping us to understand the FrameNet annotation process and 
the technical aspects of the data, and for co-organizing the FrameNet 
tutorial in which an early version of the API was introduced \citep{baker-15}. 
We also thank NLTK project leader Steven Bird, 
Mikhail Korborov, Pierpaolo Pantone, Rob Malouf, 
and anyone else who may have contributed to the release of the API by 
reviewing the code and reporting bugs; 
and anonymous reviewers for their suggestions.
}

\bibliographystyle{style/emnlp_natbib}
\bibliography{fnapi}

\end{document}